\documentclass[letter, 10 pt, conference]{ieeeconf}  
\IEEEoverridecommandlockouts                              
\usepackage{graphics}    
\usepackage{times}       
\usepackage{amsmath}     
\usepackage{amssymb}     
\usepackage{graphicx}
\usepackage{subcaption}
\usepackage{hyperref}
\usepackage{xspace}
\usepackage{xcolor}
\usepackage{color}
\usepackage{todonotes}
\usepackage[linesnumbered, ruled]{algorithm2e}
\usepackage{multirow}
\usepackage{balance}

\usepackage[font=small]{caption}

\def\secref#1{Sec.~\ref{#1}}
\def\figref#1{Fig.~\ref{#1}}
\def\tabref#1{Tab.~\ref{#1}}
\def\eqref#1{Eq.~(\ref{#1})}
\def\algref#1{Alg.~\ref{#1}}

\newcommand{\algline}[1]{$\ell$.~{\footnotesize #1}}

\newcommand{\mass}{\mathit{m}} 
\newcommand{\gravity}{\mathit{g}} 
\newcommand{\inertialVectorComponent}{\vec{\mathbf{i}}} 
\newcommand{\bodyVectorComponent}{\vec{\mathbf{b}}} 
\newcommand{\referenceInertia}{\mathnormal{F}_I} 
\newcommand{\referenceBody}{\mathnormal{F}_B} 
\newcommand{\iRb}{{^IR_B}}

\newcommand{\real}{\mathbb{R}} 
\newcommand{\SO}{\mathit{SO}}
\newcommand{\linearVelocity}{\mathit{v}}
\newcommand{\angularVelocity}{\mathit{\omega}}
\newcommand{\Trans}{\top}
\newcommand{\costFunction}{\mathcal{C}}

\newcommand{\flatoutput}{\zeta}

\usepackage{amsopn}

\newcommand{\bH}{\mathbf{H}}
\newcommand{\bI}{\mathbf{I}}

\newcommand{\bb}{\mathbf{b}}

\newcommand{\bOmega}{\mathbf{\Omega}}

\title{\LARGE \bf Non-Linear Model Predictive Control with\\Adaptive Time-Mesh Refinement}

\author{Ciro Potena$^*$ \and Bartolomeo Della Corte$^*$ \and Daniele Nardi \and Giorgio Grisetti \and Alberto Pretto 
  \thanks{$^*$~These two authors contribute equally to the work.}%
  \thanks{This work has been supported by the European Commission under the grant number H2020-ICT-644227-FLOURISH. All the authors are with Sapienza University of Rome, Department of
    Computer, Control, and Management Engineering Antonio Ruberti,
    Rome, Italy.
  }%
}

\begin{document}
\maketitle
\thispagestyle{empty}
\pagestyle{empty}

\begin{abstract}

In this paper, we present a novel solution for real-time, Non-Linear
Model Predictive Control (NMPC) exploiting a time-mesh refinement
strategy. The proposed controller formulates the Optimal Control
Problem (OCP) in terms of \textit{flat} outputs over an adaptive lattice. 
In common approximated OCP solutions, the number of discretization points
composing the lattice represents a critical upper bound for real-time applications.
The proposed NMPC-based technique refines the initially uniform time horizon by adding 
time steps with a sampling criterion that aims to reduce the discretization error.
This enables a higher accuracy in the initial part of the receding horizon, which
is more relevant to NMPC, while keeping bounded the number of discretization points.
By combining this feature with an efficient
Least Square formulation, our solver is also extremely time-efficient,
generating trajectories of multiple seconds within only a
few milliseconds. The performance of the proposed approach has been
validated in a high fidelity simulation environment, by using an
UAV platform. We also released our implementation as open
source C++ code.

\end{abstract}

\section{Introduction}
\label{sec:intro}
Nowadays, there is a strong demand for autonomous robots and unmanned
vehicles enabled with advanced motion capabilities: a mobile platform
should be able to perform fast and complex motions in a safe way, and
to quickly react to unforeseen external events. The ability to deal
with complex trajectories, fast re-planning, dynamic object tracking
and obstacle avoidance requires effective and robust motion planning
and control algorithms.

One of the most promising solution to such kind of problems is to
handle the trajectory planning and tracking together by means of a
Non-Linear Model Predictive Controller
(NMPC)\cite{grune2017nonlinear}. NMPCs formulate the problem into an
Optimal Control Problem (OCP) with a prediction horizon $T$ starting
from the current time: at each new measurement, the NMPC provides a
feasible solution and only the first control input of the provided
trajectory is actually applied to control the robot.

However, for complex trajectories that involve additional constraints
such as obstacles to avoid or objects to track, a NMPC requires to
exploit a relatively large time horizon to be effective. The typical
solution is to approximate such time horizon by means of a discrete
number $N$ of time steps: usually having a small value of $N$ leads to
a poor quality in the approximate solution while, on the other hand, a
large $N$ does not allow real-time computation, thus making NMPC
impractical in real-world applications\footnote{In a standard control
  problem within a dynamic environment, both the controller and
  planner are supposed to provide a solution in a relatively small
  amount of time (e.g. within $\sim10$ ms).}
\cite{potena2017ecmr}\cite{sheckells2016iros}.

\begin{figure}[t]
  \centering
  \includegraphics[width=\columnwidth]{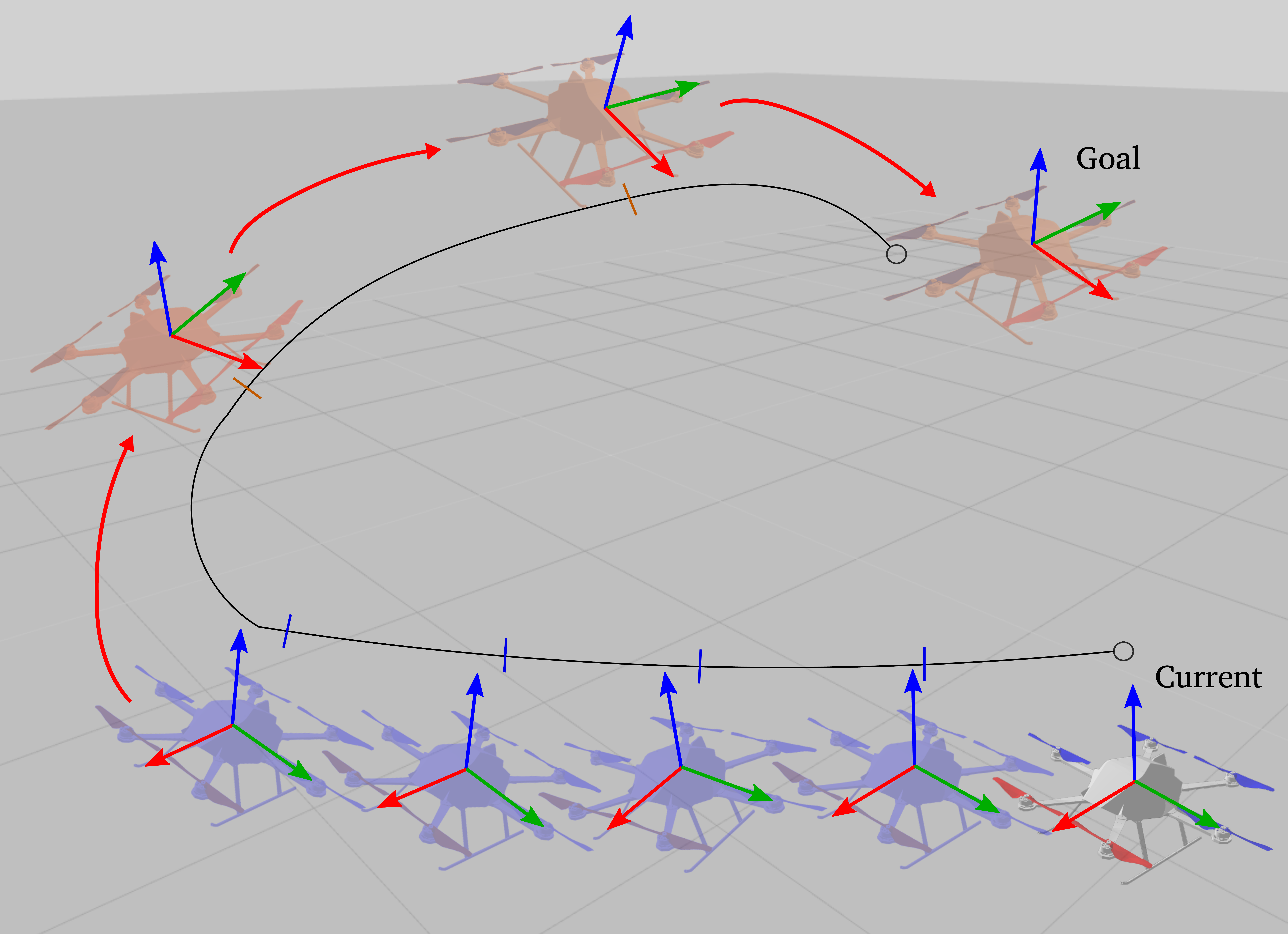}
  \caption{Illustration of the optimal trajectory computed at each
    time instant by the proposed approach. UAV frames represent
    trajectory points along the prediction horizon $T$: the blue frames
    represent the refined portion of the lattice, while the red ones
    correspond to the coarser rest of trajectory.}
  \label{fig:motivatonal}
\end{figure}

In this paper, we provide an effective solution to NMPC based
real-time control problems, by proposing a novel adaptive time-mesh
refinement strategy employed for solving the OCP, implemented inside
an open-source NMPC library released with this paper.


Our approach formulates the task of steering a robot to a
goal position, or along a desired trajectory, as a least square
problem, where a cost over the dynamics and trajectory constraints is
minimized. The optimization is performed over a discrete-time
parametrization in terms of the flat outputs \cite{fliess1995flatness} of the dynamical system,
thus decreasing the dimensionality of the OCP while reducing the
computational effort required to find the solution.

In standard OCPs, an approximate solution to the continuous problem is found by
discretizing the underlying control problem over a uniform
time lattice. The number of discretization points is one of the major factors 
influencing the accuracy of the solution and the computational time: we propose 
to mitigate this problem by focusing also on the \emph{points location}, 
so to condense more time samples where the discretization error is high.

We propose an algorithm that iteratively finds a suitable points distribution within the
time-mesh, that satisfies a discretization error criterion.
To keep the computational burden as low as possible, our approach
increases the local mesh resolution only in the initial parts of the
receding horizon, thus providing a higher accuracy in the section of trajectory that is more
relevant to the NMPC.



We provide an open source C++ implementation of the proposed solution at:

\begin{quote}
  \begin{footnotesize}
    { \url{https://bitbucket.org/gnomicSolver/gnomic}}
  \end{footnotesize}
\end{quote}

Our implementation allows to test the proposed approach on
different robotic platforms, such as Unmanned Aerial Vehicles (UAVs),
and ground robots, with both holonomic and non-holonomic
constraints.\\
We exploit this implementation in our experiments, where we evaluate 
the proposed approach inside a simulated environment using an Asctec Firefly UAV.
We proof the effectiveness of our method with a performance comparison against a
standard NMPC solution.

\subsection{Related Work}
\label{sec:related}

Recently, MPC approaches have been employed in several robotics
contexts
\cite{ginhoux2004beating}\cite{bouffard2012icra}\cite{alexis2011med}\cite{keivan2013realtime}\cite{liniger2015ocam},
thanks to on-board increased computational capabilities. In
\cite{houska2011ocam} and \cite{quirynen2015ocam} ACADO, a framework for fast
Nonlinear Model Predictive Control (NMPC), is presented;
\cite{kamel2015cca} and \cite{potena2017ecmr} use ACADO for fast
attitude control of an Unmanned Aerial Vehicle (UAV). Such a framework
has been further improved in \cite{vukov2013auto} by adding a code
generator for embedded implementation of a linear MPC, based on an
interior-point solver. Most of these methods are solving a constrained
MPC problem, which is computationally complex and thus it requires a
trade-off between time horizon and policy lag.

An alternative algorithm for solving unconstrained MPC problem is a
Sequential Linear Quadratic (SLQ) solver. One well-known variant of
SLQ called iterative Linear Quadratic Gaussian (iLQG) is presented in
\cite{todorov2005generalized} and exhaustively tested in simulation
scenarios. The feasibility of SLQ in simulation has been also
confirmed in \cite{tassa2012iros} and \cite{erez2011rss}. In recent
works \cite{garimella2015icra},\cite{farshidian2014learning} the
authors have successfully demonstrated the effectiveness of SLQ on a
real platform. However, in these projects the control is not
recomputed during run time, which limits these approaches in terms of
disturbance rejection, reaction to unperceived situations and
imperfect state estimation. These limitations can be mitigated by
solving the optimal control problem using a MPC formulation. First
implementations of such approach were shown in
\cite{erez2014receding} and \cite{koenemann2015whole}. However, in
these works time horizons are short and additional tracking
controllers are used instead of directly applying the feedforward and
feedback gains to the actuation system. This issue has been handled in
\cite{neunert2016fast}, where the authors allow SLQ-MPC to directly
act on the actuation system, hence intrinsically improving the
performance. Despite that, the size of time horizon and the number of
constraints still strongly affects the computational time required for
finding the solution. In \cite{sheckells2016iros}, the authors handle
this issue by exploiting the differential flatness property of
under-actuated robots, like Unmanned Aerial Vehicles (UAVs).  
With a reduced computational effort, the solver can run at 10 Hz, thus
limiting the vehicle's capacities to safely navigate in dynamic
environments.  These issues are confirmed in our previous work
\cite{potena2017ecmr}, where two different NMPCs have been adopted in
order to handle an Optimal Visual Servoing problem. The first serves
for estimating the optimal trajectory in terms of a target
re-projection error, while the second is employed as a high-level
controller. The complexity introduced by the target re-projection
error constraint does not allow to get a feasible solution in
real-time.  Hence, this does not enable the robot to successfully
track moving objects.

In \cite{paiva2017controlo}, the authors proposed an adaptive time–mesh
refinement algorithm that iteratively refines the uniform lattice by adding
new control points. Despite they demonstrate how the
exploitation of this algorithm allows to save up to 80\%
runtime with respect to the uniform lattice, at the best of our
knowledge, there is still no real-time implementation.

In this work we propose a Least-Square (LS) based NMPC
applied to highly dynamic, non-linear dynamic robots, employing an
adaptive time-mesh refinement and the differential flatness property.

The proposed time-mesh refinement procedure is similar to the one presented in 
\cite{paiva2017controlo}, but with some important differences induced 
by the real-time application context: (i) a maximum number of refinement 
iteration; (ii) the usage of a discretization error as refinement criteria;
(iii) the focus on the initial part of the coarse optimal trajectory.

By exploiting those properties in our efficient LS implementation, we
are able to solve the NMPC problem for time horizons of multiple
seconds in only a few milliseconds, obtaining unmatched performance by
similar state-of-the-art algorithms. From exhaustive experiments in
simulated environments, we demonstrate the capabilities of the
approach in navigation tasks that leverage the full system dynamics.

\subsection{Contributions}
\label{sec:contribution}

The method proposed in this paper differs from previous
works in the following aspects:

(i) it uses a novel adaptive time-mesh refinement algorithm that
allows to reduce the computational burden, while keeping an adequate
level of precision in the optimal solution;
(ii) it exploits the differential flatness property of under-actuated
vehicles, enabling to reduce the problem complexity;
(iii) its generality allows for ease adaptation and extension to
heterogeneous robotic platforms.

Moreover, we released a C++ open-source implementation for different robotic platforms.

Our claims are backed up through the experimental evaluation.
\section{Adaptive Finite Horizon Optimal Control}
\label{sec:approach}

\subsection{Problem Statement}
The goal of this work is to generate an optimal trajectory along with
the set of control inputs to enable a robot to closely track it.
We formulate this non-linear OCP in a NMPC fashion over the finite 
time horizon $T$.  We assume a general non-linear time-step rule of 
the dynamics as

\begin{equation}
\label{eq:nonlinear_model}
x_{k+1} = f(x_k,u_k) 
\end{equation}

where $x_k$ and $u_k$ denote the state vector and the control input
vector at the time $t_k$, respectively. $f(x_k,u_k)$ represents the
non-linear dynamic model. It is assumed to be differentiable with
respect to the state $x_k$ and control input $u_k$, and it maps the
state and the input vector in the subsequent time $t_{k+1}$. The
time-step of the dynamics is given by $T/N$, so that each $x_k$ occurs
at time $t_i = it_f/N$, where $N$ is the number of trajectory points
used for finding the optimal solution and $t_f$ indicates the desired
final time. The goal is to find an optimal time-varying feedback and
feedforward control law of the form:

\begin{equation}
u(x_k,k) = ufb(x_k) + uff(x_k)
\end{equation}

where $uff(x_k)$ is the feedforward term while $ufb(x_k)$ represents the
feedback control term. The optimal time-varying control law is found
by iteratively finding an optimal solution for minimizing a cost
function $\costFunction$.

\subsection{Overview of the Control Framework}
\label{sec:overview}

The cost function $\costFunction$ is composed by a set of constraints
over $N$ time steps of the state vector $x_{0:N} = \{x_0, x_1, \dots,
x_N\}$ along with the corresponding input controls $u_{0:N-1} = \{u_0,
u_1, \dots, u_{N-1}\}$.  Defining $x_f$ as the desired goal state, the
cost function can be formulated in a standard optimal control form as:

\begin{equation}
\hat{\costFunction} = \frac{1}{2} \sum_{k=0}^{N-1} \left( \| x_k - x_f \|_Q^2 + \| u_k \|_R^2 \right)
\end{equation}
\noindent

where $\| \cdot \|_\Omega^2$ stands for the $\Omega$-norm, $Q\geq0$ is
the matrix that penalizes the distance to the goal state, and $R>0$ is
the matrix weighting the control inputs.  Moreover, in order to enable the
solver to take advantage of the vehicle dynamic model, we add a
continuity constraint between subsequent states as:

\begin{equation}
\label{eq:cost_function_with_albrd}
  \costFunction = \hat{\costFunction} + \frac{1}{2} \sum_{k=1}^{N-1} \left( \| x_{k+1} - f(x_{k}, u_{k}) \|_{A_l}^2 \right)
\end{equation}
\noindent

where the $A_l$ matrix weights the single state component continuity,
with $A_l > 0$. Intuitively, temporally adjacent states are forced to
attain the system dynamics.

At this points, we adopt a simplifying working assumption in order to 
reduce the computational cost of the optimization procedure
employed for minimizing the cost function stated in
\eqref{eq:cost_function_with_albrd} and, therefore to handle non-linear
constraints in an on-line implementation: the assumption consists of 
restricting our attention to non-linear
differential flat systems \cite{fliess1995flatness}. As well-known, for
such kind of systems we can find a set of outputs $\flatoutput \in
\mathbb{R}^m$, named \textit{flat}, of the form
\begin{equation}
 \flatoutput = h(x,u,\dot{u},\cdots,u^{(r)})
\end{equation}
such that there exist two functions $\psi_k$ and $\phi_k$ for which 
the state and the input can be expressed in terms of flat states 
and a finite number of their derivatives
\begin{align}
& x_k = \psi_k(\flatoutput, \dot{\flatoutput},\cdots,\flatoutput^{(c)}) \\
& u_k = \phi_k(\flatoutput, \dot{\flatoutput},\cdots,\flatoutput^{(c)}) 
\end{align}
The formulation of the OCP in terms of \textit{flat} states allows for a substantial dimensionality 
reduction of the problem, and consequently to a saving in terms of computational cost.
The cost function may consequently be rewritten using the \textit{flat} outputs as
\begin{equation}
  \hat{\costFunction}(\flatoutput) = \frac{1}{2} \sum_{k=0}^{N-1} \left( \| \underbrace{\psi_k(\flatoutput) - x_f}_{\nu_k} \|_Q^2 + \| \phi_k(\flatoutput) \|_R^2 \right)
\end{equation}
and, by adding also the continuity constraints, the whole cost function is calculated as
\begin{multline}
\label{eq:flat_err_function}
 \costFunction(\flatoutput) =  \hat{\costFunction_0}(\flatoutput) + \frac{1}{2} \sum_{k=1}^{N-1} \Bigg( \| \nu_k(\flatoutput) \|_Q^2 + \| \phi_k(\flatoutput) \|_R^2 +\\+ \| \underbrace{{\psi_{k+1}}(\flatoutput) - f(\psi_{k}(\flatoutput), \phi_{k}(\flatoutput))}_{\gamma_k} \|_{A_l}^2 \Bigg)
\end{multline}
To find the solution for the cost function in 
\eqref{eq:flat_err_function}, we adopt a well-established numerical
method for solving optimal control problems, namely direct multiple
shooting \cite{betts1998survey}. In direct multiple shooting, the
whole trajectory is parametrized by finite number of \textit{flat}
outputs $\flatoutput \in \mathbb{R}^{Nm}$. Hence, by stacking all the error
components that compose the cost function~\eqref{eq:flat_err_function},
we obtain the error function $e(\flatoutput)$ as

\begin{equation}
  \label{eq:error_function}
e(\flatoutput) =  
\begin{bmatrix}
   \nu_0(\flatoutput)\\
   \phi_0(\flatoutput)\\
   \gamma_0(\flatoutput)\\
   \vdots \\
   \nu_{N-1}(\flatoutput)\\
   \phi_{N-1}(\flatoutput)\\
   \gamma_{N-1}(\flatoutput)\\
\end{bmatrix}
\end{equation}

We minimize the error function in \eqref{eq:error_function} by adopting a
Least-Square iterative procedure, where the trajectory is iteratively
updated as
\begin{equation}
  \flatoutput \leftarrow \flatoutput \boxplus \delta\flatoutput
\end{equation}
where the $\boxplus$ operator performs the variable update, while taking into 
account the specific composition of the \textit{flat} state \cite{smith1986representation}.
The update vector $\delta\flatoutput$ is found by solving a linear system of
the form $\bH \delta\flatoutput = \bb$ with the terms $\bH$ and $\bb$  given by
\begin{eqnarray}
  \bH &=&  J(\flatoutput)^\Trans \bOmega J(\flatoutput)\\
  \bb &=&   J(\flatoutput)^\Trans \bOmega e(\flatoutput)
\end{eqnarray}
where $J(\flatoutput) = \partial e(\flatoutput)/\partial\flatoutput$. To limit the 
magnitude of the perturbation between iterations and thus, enforce a 
smoother convergence, we solve a damped linear system of the form
\begin{equation}
\label{eq:linsys}
\left (\bH + \lambda \bI \right) \delta\flatoutput = \bb.
\end{equation}
\section{ADAPTIVE TIME-MESH REFINEMENT}

In non linear OCPs, the choice of the number of trajectory points $N$
is a major factor affecting the computational cost that is required
to get a solution and also the accuracy of the solution itself.
Hence, our goal is to effectively arrange the trajectory points along
the time horizon $T$.

In this work we address the trajectory points displacement by means of
an adaptive time-mesh refinement strategy. 
We start finding an initial solution by solving the OCP, as described in 
\secref{sec:approach}, with a coarse lattice. Our method looks for portions 
of the horizon $T$ that need to be refined: we sample here new keypoints 
with a fine granularity.


In particular, since the NMPC employs a receding horizon strategy --
where only the first trajectory point is actually used for the system 
actuation -- we focus the resampling procedure on the initial part of 
the horizon.

This allows us to achieve adequate tracking performance even with a
minimum amount of trajectory points $N$. As a consequence, solving more complex
OCPs with longer time horizons $T$ or additional constraints
(\textit{e.g.} obstacles to avoid, objects to track, \textit{etc.})
can be handled in real-time.


More formally, once the initial problem has been solved by employing
the procedure described in \secref{sec:overview} over an uniform
time-mesh, the solution is iteratively processed.  
As reported in
\algref{alg:time_mesh_refinement}, at each iteration, the time-mesh
refinement performs the following steps: (i) it performs a
discretization error check  (\algline{2} and \algline{12}), which allows detecting where  new trajectory points are required; (ii) it then adds new points by
interpolating them between the adjacent ones (\algline{7-8}); (iii) finally, it
transcribes the sub-problem into an OCP and solves it (\algline{10-11}).
In the following, we discuss these steps in more
detail.

\subsection{Discretization Error Check}

In order to proceed with the time-mesh refinement strategy, we have to
define a refinement and a stopping criteria. In this work, we consider
as main refinement criterion the discretization error between the
\textit{flat} variables. The discretization error, at each lattice point,
is computed as the difference between the current state and a higher
order approximation of the solution trough the non-linear time-step
rule of the dynamics of \eqref{eq:nonlinear_model}. More specifically,
starting from the $\flatoutput_{k-1}$ node, we perform a finer
integration of the dynamics with respect to the one employed during
the OCP solution, obtaining $\hat{\flatoutput}_k$. Hence, the
discretization error is computed as
\begin{equation}
\epsilon_{\flatoutput_k} = \| \hat{\flatoutput}_k - \flatoutput_k \| 
\end{equation}
where $\|\cdot\|$ stands as the squared norm of two vectors.  As
stopping criterion we consider a threshold on the discretization error
$\epsilon_{\flatoutput_k}$. 

\subsection{Trajectory Points Interpolation}

Once the discretization error has been obtained, the time-mesh
refinement proceeds by adding new trajectory nodes where required.  In
order to obtain a smooth interpolation that preserves the
\textit{flat} states differentiability, we use the cubic Hermite
interpolation. More formally, let $\flatoutput_i$ and $\flatoutput_j$ 
be two adjacent \textit{flat} states in the lattice, the interpolated 
\textit{flat} state in the unit time interval $[0,1]$ is computed as:
\begin{multline}
  \flatoutput_k = \flatoutput_i (2t^3 - 3t^2 + 1) + \flatoutput_i^{'} (t^3 - 2t^2 + t)\\+ \flatoutput_j (3t^2 -2t^3) +\flatoutput_j^{'}(t^3 - t^2)
\end{multline}
where $t\in[0,1]$ is the interpolation point, and $\flatoutput^{'}$
denotes the first-order \textit{flat} variable derivative.

\subsection{Local Optimization Procedure}

The refinement procedure adds more node points to the initial
portion of the trajectory where the discretization error is higher, 
so to obtain an improved solution.

As a consequence, the computational time increases. To avoid this
issue, when progressively going from a coarse mesh to a refined one,
the error function in \eqref{eq:error_function} is scaled only to the initial
part of the horizon. More specifically, let $N_{tm} = N_i + N_{add}$ be the number of initial
trajectory points that are going to be refined, where $N_i < N$ is an user defined parameter 
representing the number of points that belong to an initial section of the trajectory,
and $N_{add}$ is the number of points added at each 
iteration of the refinement algorithm. At each iteration, the refinement procedure transcribes these points in an OCP and re-optimizes
only the mesh sub-intervals that belong to $\flatoutput_{0:N_{tm}} =
\{\flatoutput_0,\cdots,\flatoutput_{N_{tm}}\}$, while keeping
unchanged the remaining part of the trajectory.
\begin{algorithm}
  \small
  \KwData{Cost function $\costFunction$, dynamics $f(x(k),u(k))$, OCP solution $\zeta_{0:N} = \{\zeta_0,\cdots,\zeta_N\}$, trajectory points to refine $N_{tm}$.}
  \KwResult{Refined trajectory $\zeta_{0:N+N_{add}}$}
  
  scale the OCP problem: $\zeta_{0:N_{tm}} = \{\zeta_0,\cdots,\zeta_{N_{tm}}\}$; \\
  discretization error computation for each lattice point $\epsilon_{\zeta_{1:N_{tm}}}$; \\
  $iter = 0$; 
  
  \While{$iter < max_{iter}$}
        {
          \ForEach{$ \zeta_i \in \zeta_{1:N_{tm}}$}
                  {
	                  \uIf{ $\epsilon_{\zeta} > err_{trs}$ }
	                      {
	                        $\flatoutput_k = Interpolate(\flatoutput_i, \flatoutput_{i+1})$; \\
	                        Add $\flatoutput_k$ to $\zeta_{0:N_{tm}}$; \\
	                        $N_{tm} \leftarrow N_{tm}+1 $; \\
	                      }
                        transcribe the scaled OCP; \\
                        apply the Least-Square solver; \\
                        estimate the discretization error $\epsilon_{\zeta_{1:N_{tm}}}$; \\
                        $iter \leftarrow iter+1$; \\
                  }
        }
        \caption{Time-Mesh Refinement Algorithm.}
        \label{alg:time_mesh_refinement}
\end{algorithm}

\begin{table*}[ht]        
     		\centering
     		\caption{Errors and Control Inputs Statistics for the Pose Regulation Experiment}
     		\label{tab:pose_regulation_rmse}
     		\begin{tabular}{ cccccccccc}
		  & $N$  & $TM ref $   & $Runtime [ms]$ & $err_{trans} [m]$ & $err_{rot}[rad]$ & $Roll_{ref}[rad]$ & $Pitch_{ref}[rad]$ & $Yaw Rate[rad/s]$ & $Thrust[Nm]$\\
		  \cline{2-10}
		  \rule{0pt}{3ex} & 100  &             & 42          & 0.0537 	  	 & 0.0435	    & 0.0196       	& 0.0196    	     & 0.3305            & 15.2450 \\ 
		  & 50   &             & 10          & 0.0613 	  	 & 0.0543	    & 0.0184       	& 0.0210             & 0.2906     	 & 15.1695 \\ 
		  & 50   & \checkmark  & 10.2        & 0.0608 	  	 & 0.0521	    & 0.0191       	& 0.0213             & 0.2863     	 & 15.2237 \\ 
		  & 20   &    	       & 1           & 0.0724 	  	 & 0.0598	    & 0.0195       	& 0.0259             & 0.3146     	 & 15.1554 \\ 
		  & 20   & \checkmark  & 1.2         & 0.0703 	  	 & 0.0553	    & 0.0193       	& 0.0224             & 0.2929     	 & 15.2491 \\ 
		  & 10   &             & 0.5         & 0.1096 	  	 & 0.0735	    & 0.0207       	& 0.0257             & 0.2964     	 & 15.1352 \\
		  & 10   & \checkmark  & 0.7         & 0.0823 	  	 & 0.0642	    & 0.0197       	& 0.0243             & 0.3321     	 & 15.1934 \\ 
		  & 5    &             & 0.2         & fail   	  	 & fail	  	    & fail         	& fail               & fail       	 & fail    \\
		  & 5    & \checkmark  & 0.4         & 0.1032 	  	 & 0.0667	    & 0.0219       	& 0.0234             & 0.3219     	 & 15.2154 \\\cline{2-10}\vspace{.1cm}
     		\end{tabular}
\end{table*}

\section{Experimental Evaluation}
\label{sec:exp}

We tested the proposed approach in a simulated environment by using
the RotorS simulator \cite{furrer2016springer} and a multirotor model.
The mapping between the high-level control input and the propeller
velocities is done by a low-level PD controller that aims to resemble
the low-level controller that runs on a real multirotor.

The evaluation presented here is designed to support the claims made in
the introduction. We performed two kind of experiments, namely
\textit{pose regulation} and \textit{trajectory tracking}.

We provide a direct comparison between a standard NMPC implementation
and the one presented in this paper, \textit{i.e.} by using a
time-mesh refinement strategy.

We formulate the OCP by composing each flat state of the UAV simulated model
as follows:
\begin{equation}
  \begin{array}{ccc}
    \flatoutput = (p_1, p_2, p_3, \gamma)
  \end{array}
\end{equation}
where $p_i$ is the translation in the world coordinate reference system along the $i^{th}$ axis, and
$\gamma$ represents the \textit{yaw} angle. For more detail about the dynamical model and the flat model,
please refer to Appendix \ref{sec:dynamic_flat_model}.

\subsection{Pose Regulation}
\label{ex:pose-regulation}
We disegned the following pose regulation experiment to prove
the accuracy and the robustness of the proposed approach.
In all the regulation experiments we set the desired state to be:
\begin{equation}
\begin{array}{ccc}
t_f = 2,  & \flatoutput = (2,2,1,1.57), &\dot{\flatoutput} = (0,0,0,0).
\end{array}
\nonumber
\end{equation}
and the time-mesh refinement parameters to be $err_{trs}~=~10^{-5},\quad max_{iter}=2.$

The pose regulation tasks are performed while varying the bins number
$N$. Our goal is to show how the control accuracy deteriorates while
going from a finer lattice to a coarser one.  To this end, we started
with $N = 200$, that we used as a reference for the other bin setups.
We decreased the bins amount up to $N = 5$ measuring the translational and rotational
RMSE with respect to the reference trajectory, for both the standard
NMPC case and our approach. We report the results of this comparison
in \tabref{tab:pose_regulation_rmse}, along with the computational
time required for solving the OCP and the control inputs average.

\begin{figure*}[ht!]
\centering
  \begin{subfigure}{.245\linewidth}
    \centering
    \includegraphics[width=\columnwidth,height=30mm]{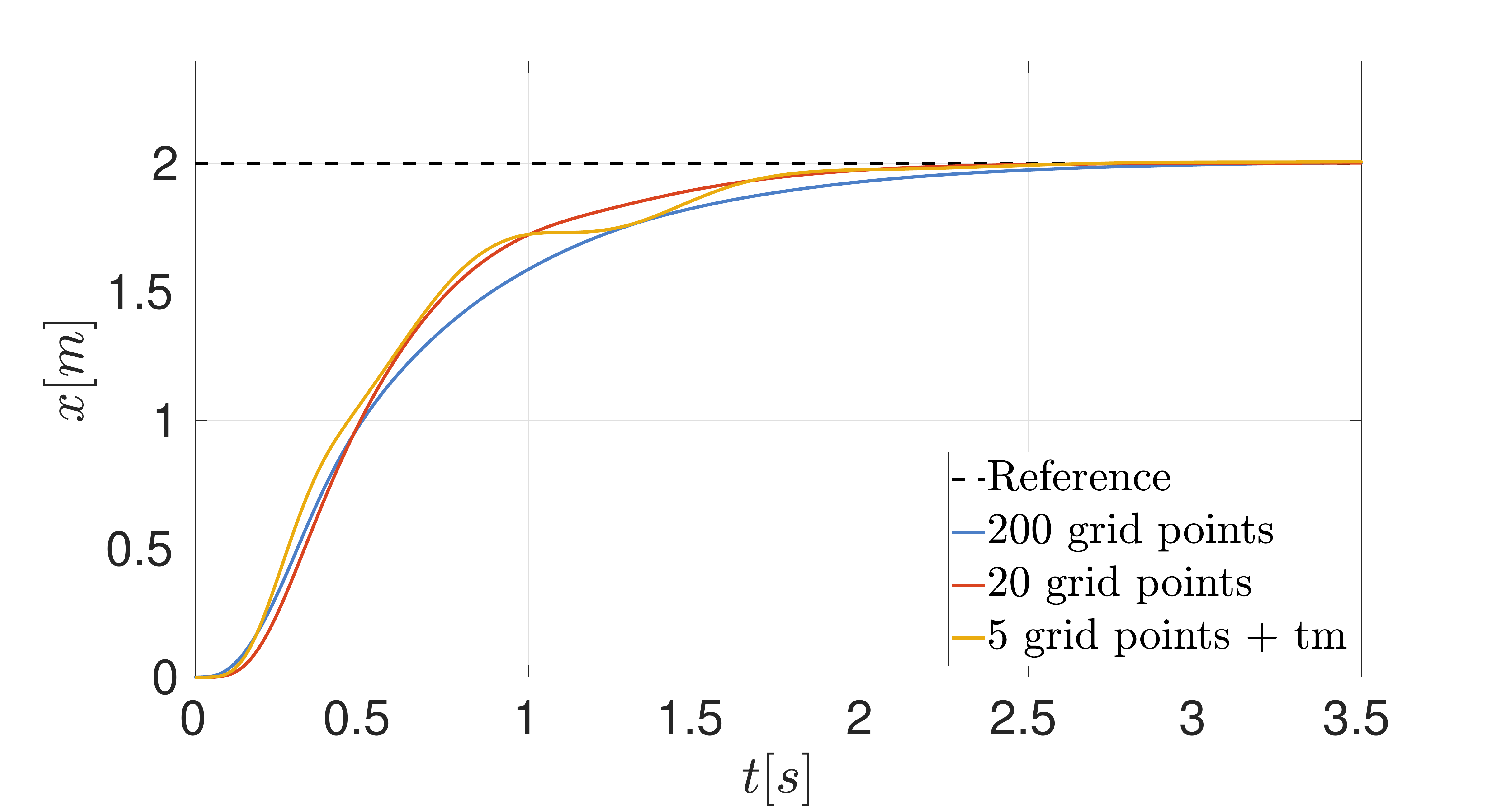}
    \caption{\textit{x-coord} error}
    \label{fig:px_way}
  \end{subfigure}
  \begin{subfigure}{.245\linewidth}
    \centering
    \includegraphics[width=\columnwidth,height=30mm]{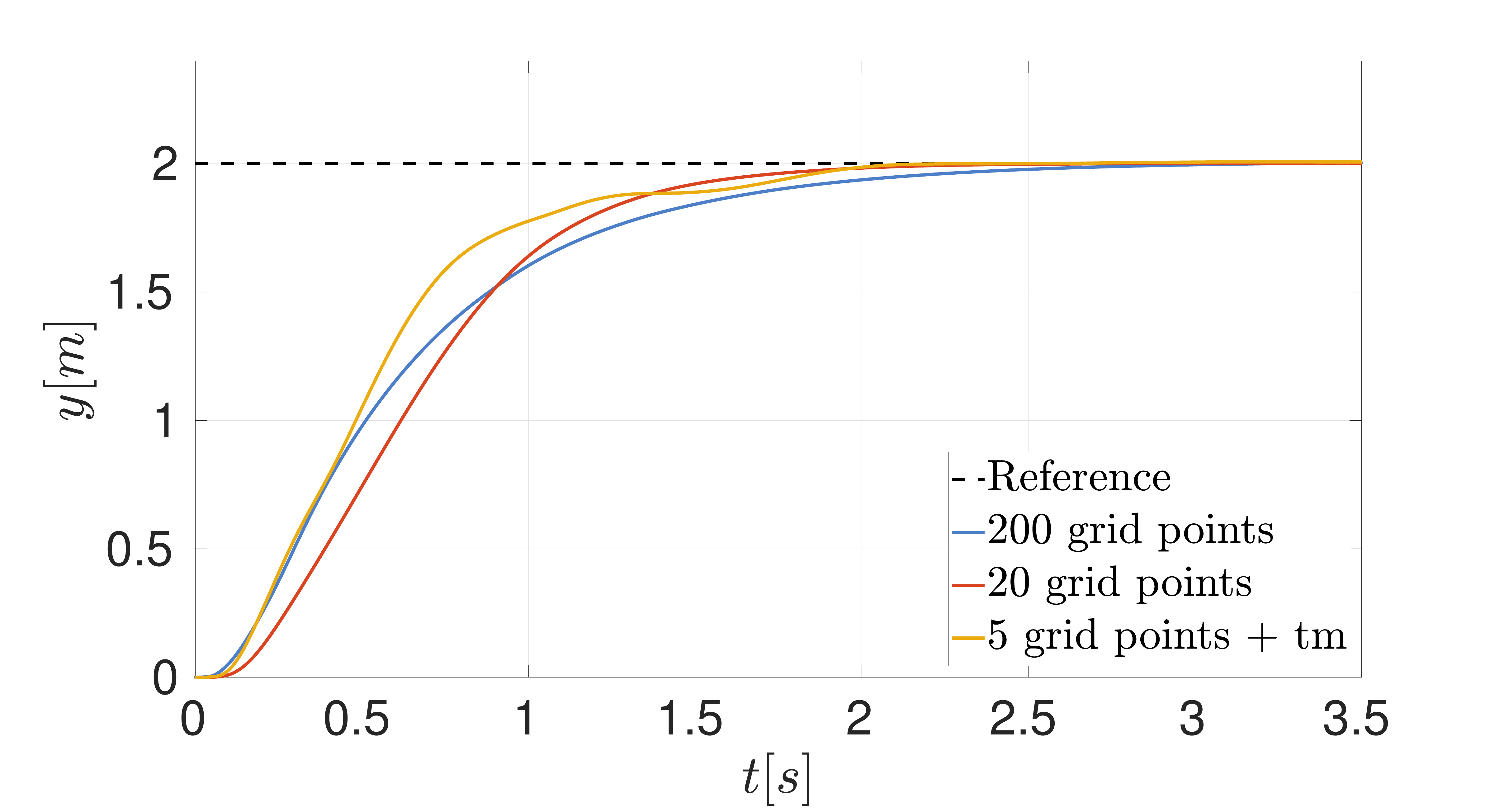}
    \caption{\textit{y-coord} error}
    \label{fig:py_way}
  \end{subfigure}
  \begin{subfigure}{.245\linewidth}
    \centering
    \includegraphics[width=\columnwidth,height=30mm]{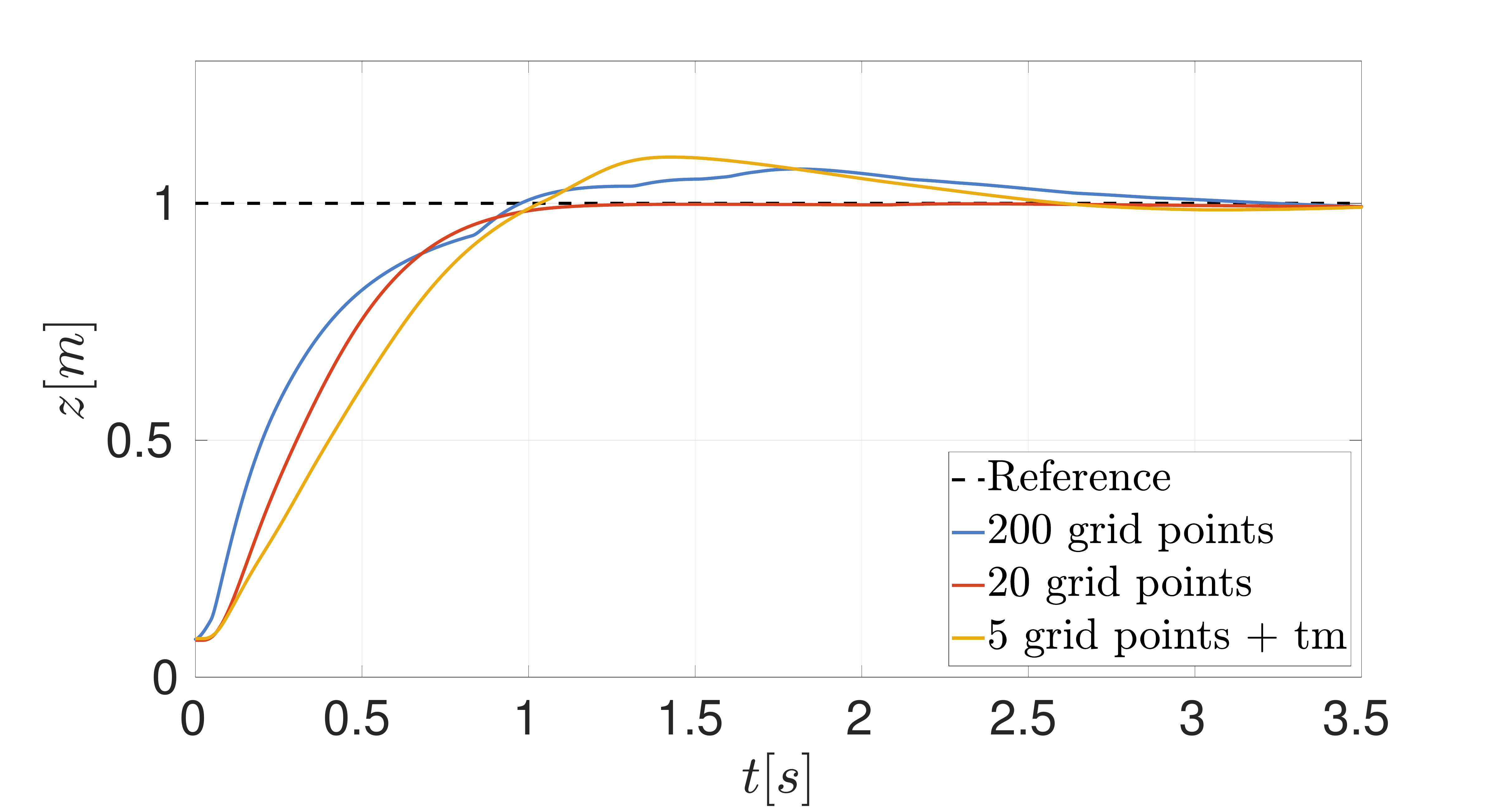}
    \caption{\textit{z-coord} error}
    \label{fig:pz_way}
  \end{subfigure}
  \begin{subfigure}{.245\linewidth}
    \centering
    \includegraphics[width=\columnwidth,height=30mm]{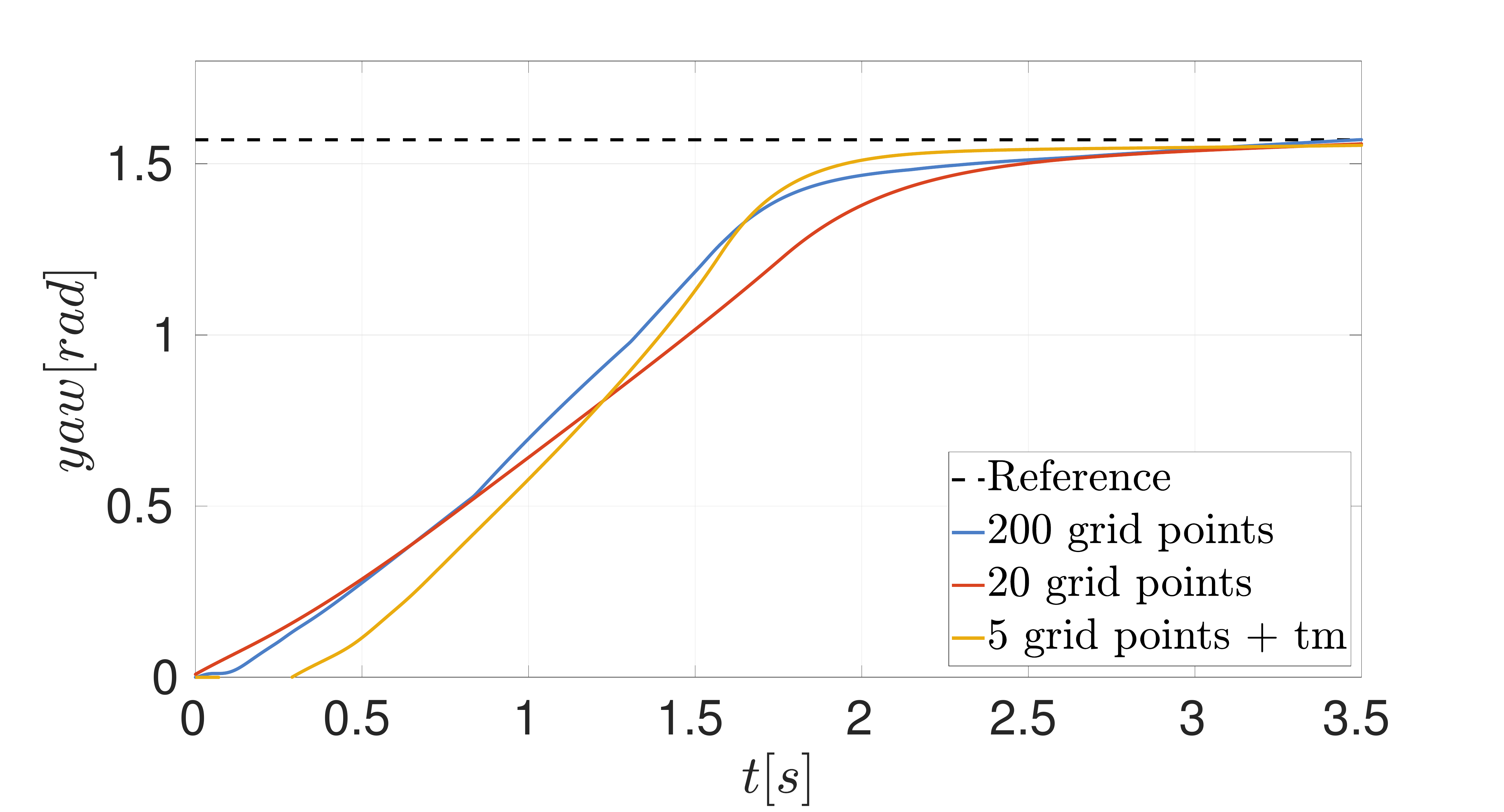}
    \caption{\textit{yaw} error}
    \label{fig:yaw_way}
  \end{subfigure}
  \caption{Pose Error}
  \label{fig:pose-error}
\end{figure*}

The advantages of using a time-mesh refinement strategy are twofold. From one side, when
using a coarse lattice, as in the case of $N=20$ and $N=10$, the
refined solution provides lower errors, with a negligible increment of
computational time.  On the other hand, it intrinsically increases the
robusteness by adaptively adding bins in the trajectory where
needed, thus avoiding failures such as the one registerd with $N = 5$
in the standard NMPC formulation. \figref{fig:pose-error} directly
compares the convergence with the reference trajectory ($N=200$) and
the ones recorded while using $N=20$ and $N=5$ with the time-mesh refinement.

\subsection{Trajectory Tracking}

To prove the effectiveness of the proposed time-mesh refinement in a tracking 
scenario, we command the UAV platform to track a challenging Lemniscate 
trajectory, defined by the following expression:
\begin{align}
  p_1(t) &= 2 sin(t/2)\nonumber\\
  p_2(t) &= 2 sin(t/2) cos(t/2)\nonumber\\
  p_3(t) &= sin( t + 5)/3\nonumber\\
  \gamma(t) &= sin(t/8)\nonumber
\end{align}
We set $t_f = 0.5s$ and $N=5$ for both the standard and the refined setup, while
we set $err_{trs}$ and $max_{iter}$ with the same values used in the pose regulation 
section \ref{ex:pose-regulation}.
We report the results of the tracking experiments in \figref{fig:lemniscate_noTM} 
and \figref{fig:lemniscate_TM}. As expected, the use of a time-mesh
refinement strategy allows for a more accurate tracking since the OCP
problem is solved over an adapted lattice.

\subsection{Runtime}
We recorded the time needed to solve the NMPC problem, as described in
\secref{sec:approach}.  We performed all the presented experiments on a
laptop computer equipped with a i7-5700HQ CPU with 2.70 GHz.  Our
software runs on a single core and in a single thread.

\begin{figure}
  \centering
  \includegraphics[width=.85\columnwidth]{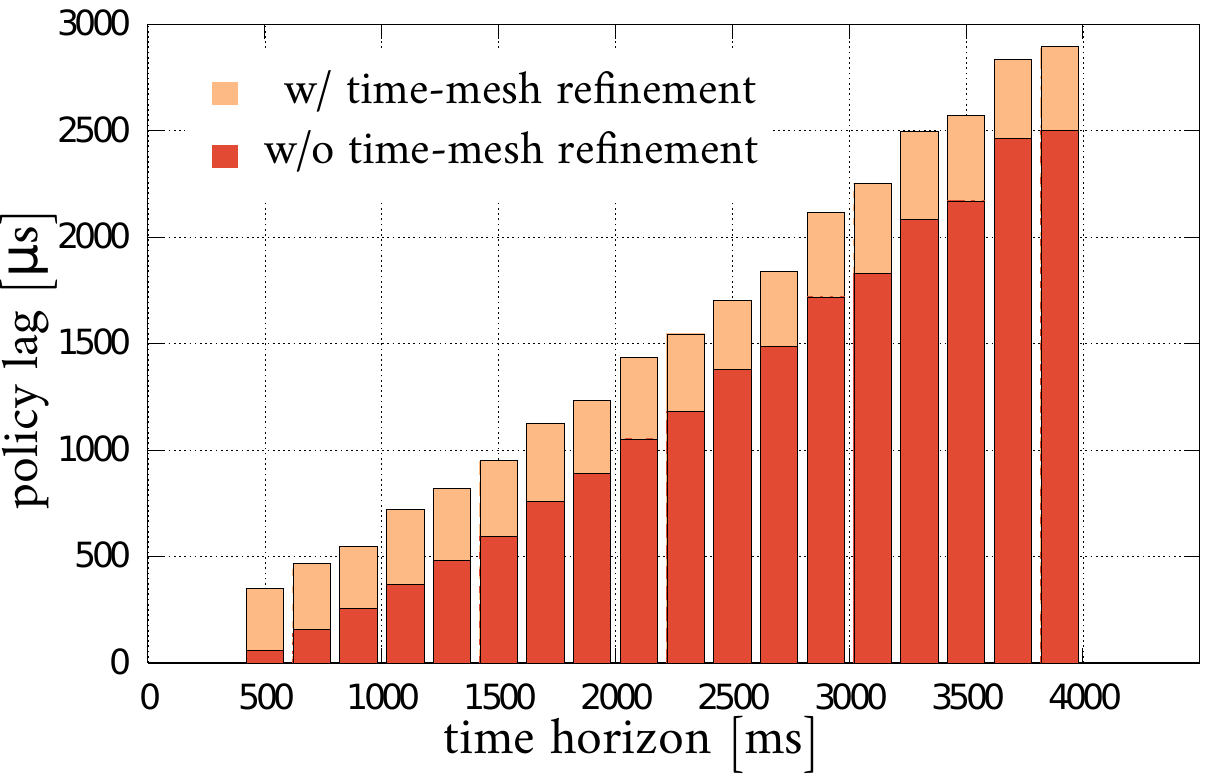}
  \caption{Average runtime over 4000 planned trajectories with horizon
    up to $4s$, with an UAV model.}
  \label{fig:computational-time-horizon}
\end{figure}

To reduce the noise in the measurements, we collect the computational
times for time horizons between $500ms$ and $4000ms$, using bins of
$200ms$. For each configuration of the solver, we compute the average
computational time over 4000 planned trajectories with an UAV, and
report these values in \figref{fig:computational-time-horizon}.

As shown in \figref{fig:computational-time-horizon}, while increasing the time 
horizon, and consequently the number of bins,
the computational time grows almost linearly. The computational overhead of the time-mesh refinement,
in this sense, does not affect such a cost, by adding a constant time to the
total computation.

\begin{figure}
  \centering
  \includegraphics[width=.85\columnwidth]{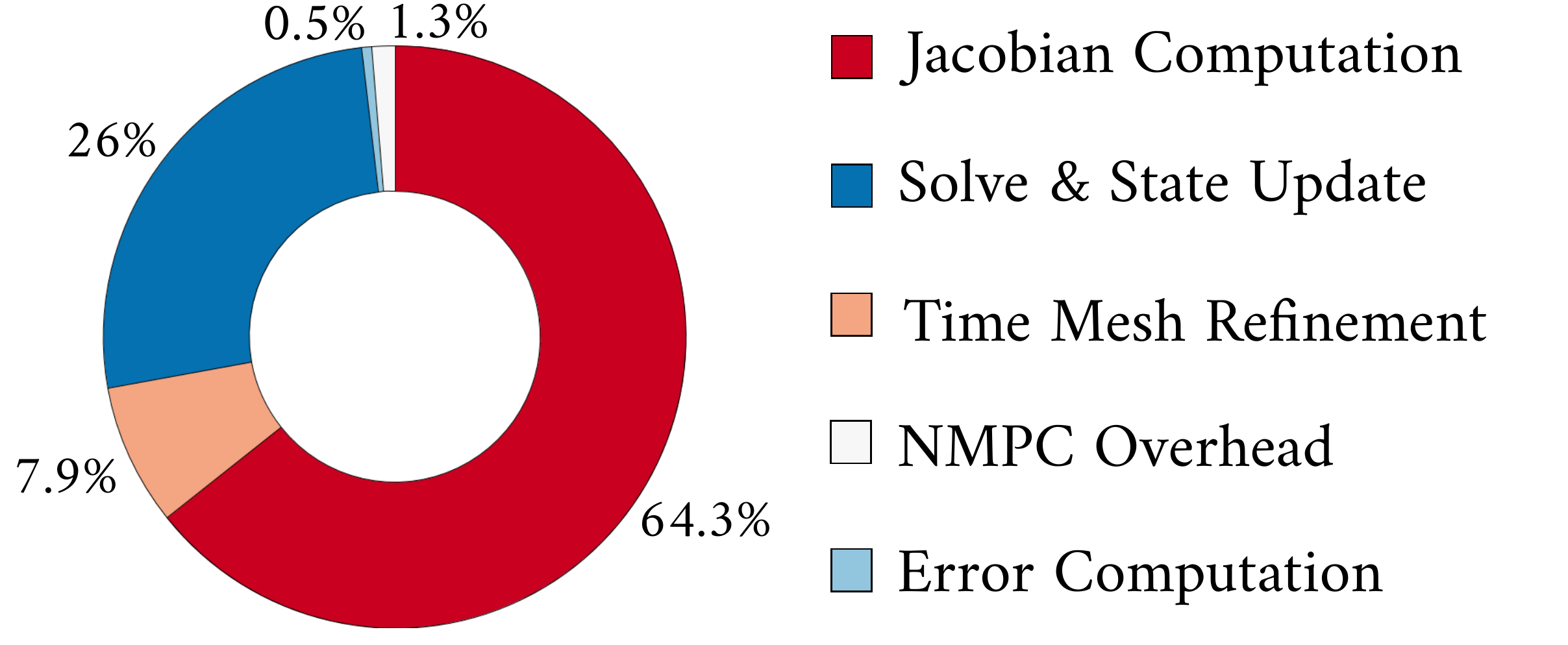}
  \caption{Runtime percentage of the NMPC algorithm with time-mesh refinement in case of $t_f = 2s$ with bins each $200ms$.}
  \label{fig:computational-pie}
\end{figure}

\figref{fig:computational-pie} shows the runtime percentage in case
of $t_f = 2s$ with bins each $200ms$, where the time-mesh refinement slice 
includes the computational time involved in all the different mesh refinement steps.
Note that the time-mesh refinement portion has constant time consumption. 
Thus its percentage value is indicative of this particular setup only. It is also 
noteworthy to highlight how most of the runtime is absorbed by the Jacobian calculation,
being computed in a fully numerical manner.

\begin{figure}[h!]
  \centering
  \begin{subfigure}{.5\columnwidth}
    \includegraphics[width=\columnwidth]{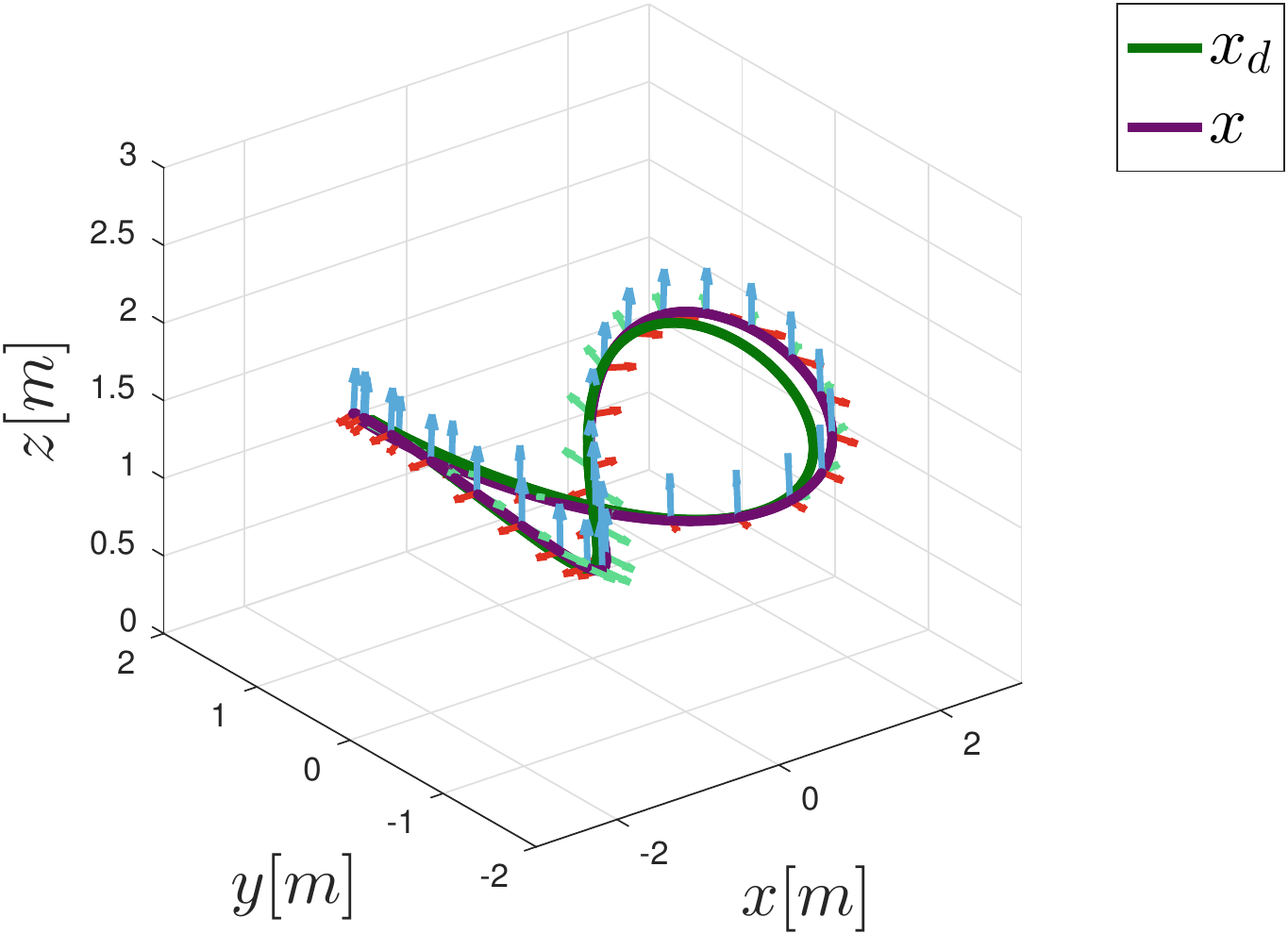}
    \caption{Trajectory Tracking.}
    \label{fig:lemniscate_noTM}
  \end{subfigure}
  \begin{subfigure}{.45\columnwidth}
    \includegraphics[width=\columnwidth]{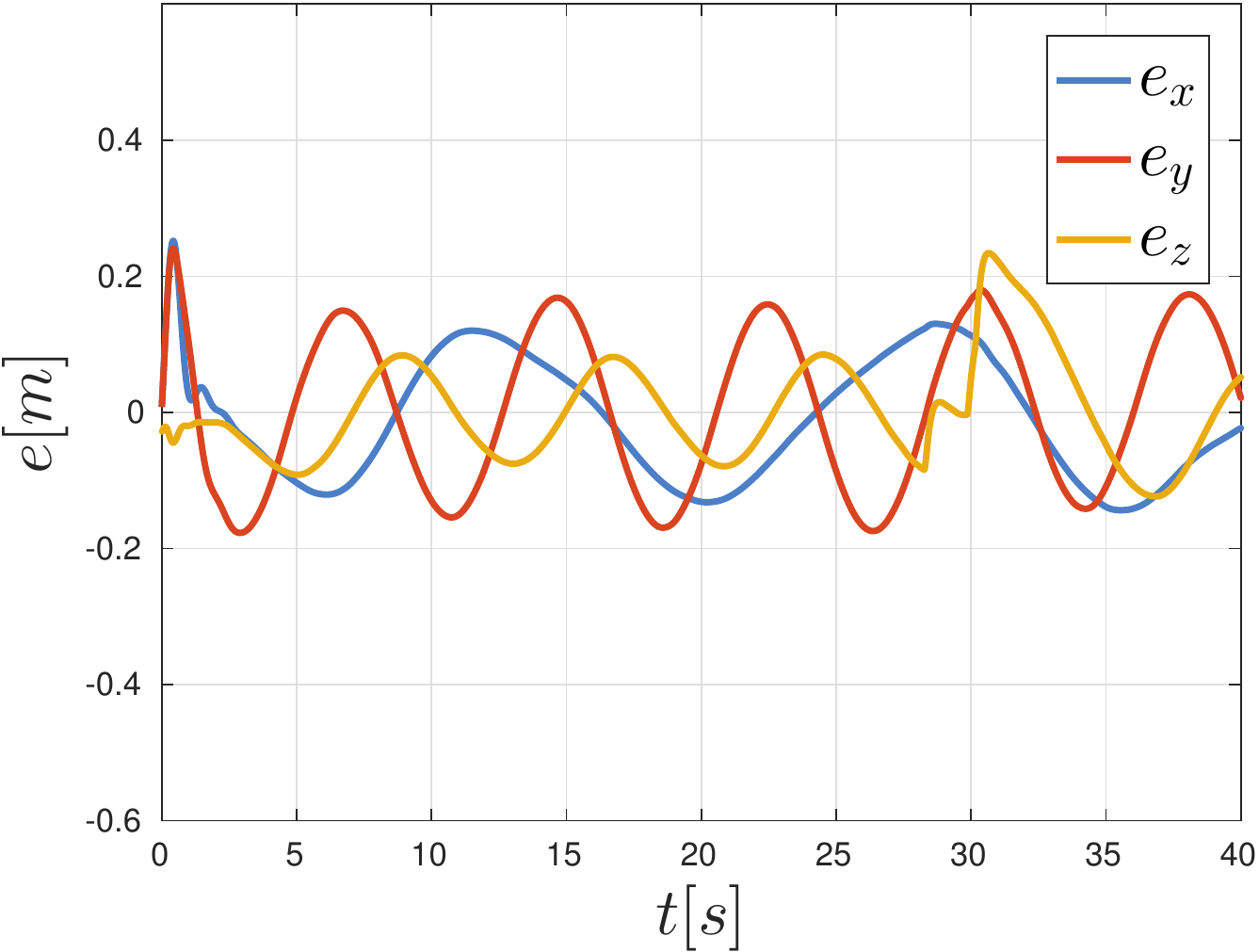}
    \caption{Position Error.}
    \label{fig:error_lemniscate_noTM}
  \end{subfigure}
  \caption{Lemniscate trajectory tracked with standard approach.}
  \label{fig:lemniscate_noTM}
\end{figure}

\begin{figure}[h!]
  \centering
  \begin{subfigure}{.5\columnwidth}
    \includegraphics[width=\columnwidth]{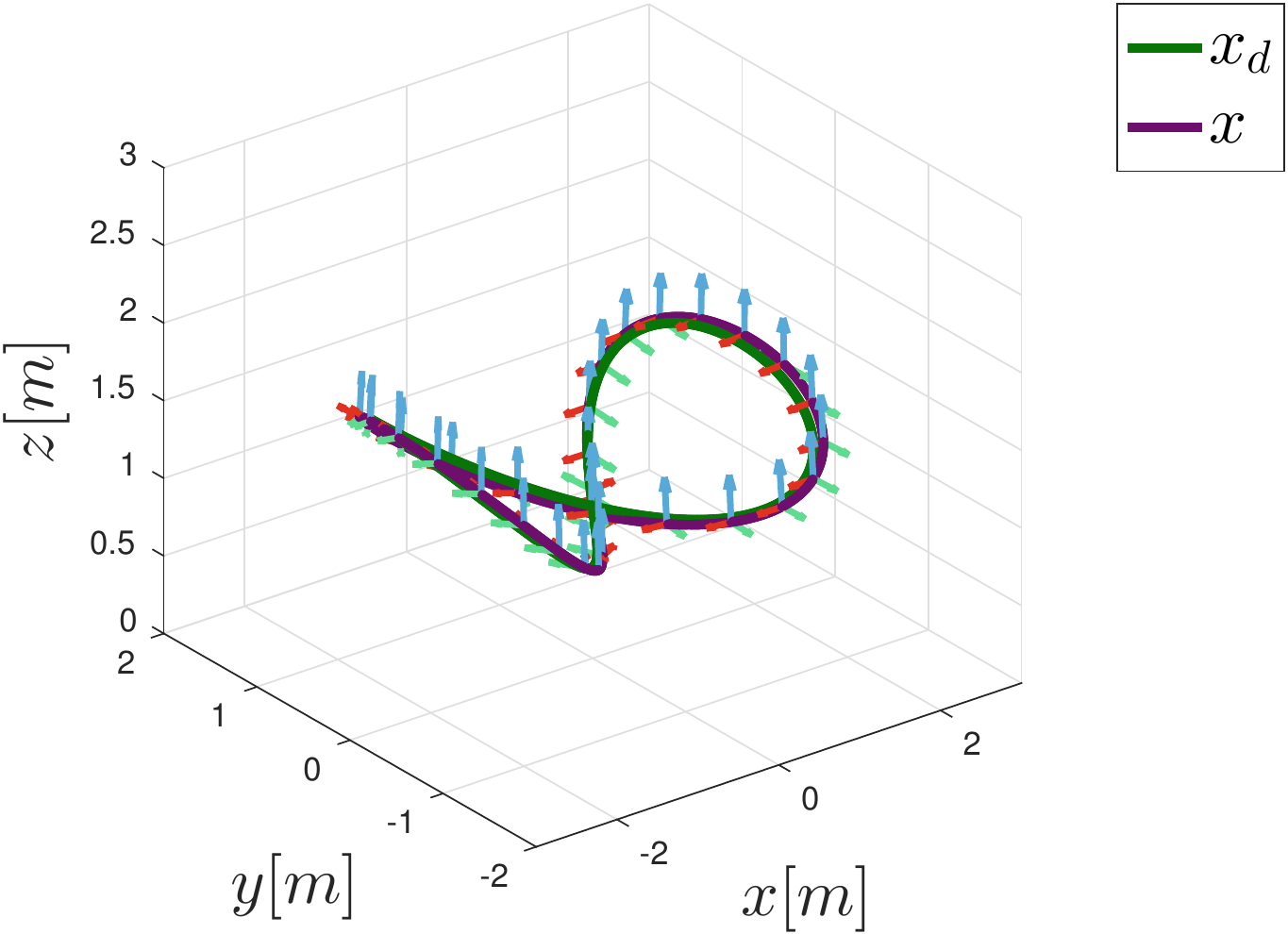}
    \caption{Trajectory Tracking.}
    \label{fig:lemniscate_TM}
  \end{subfigure}
  \begin{subfigure}{.45\columnwidth}
    \includegraphics[width=\columnwidth]{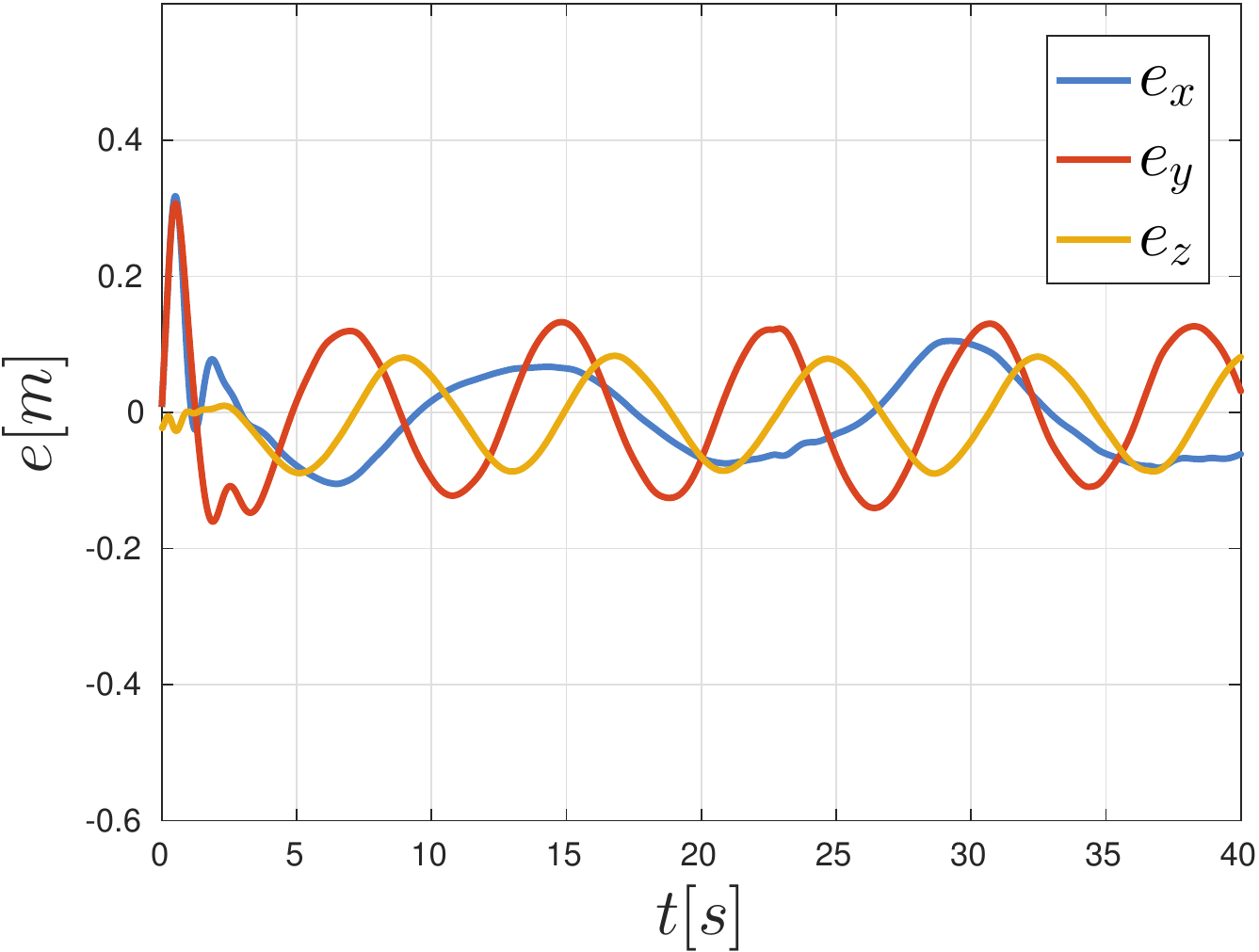}
    \caption{Position Error.}
    \label{fig:error_lemniscate_TM}
  \end{subfigure}
  \caption{Lemniscate trajectory tracked with time-mesh refinement.}
  \label{fig:lemniscate_TM}
\end{figure}

\section{Conclusion}
\label{sec:conclusion}

In this paper, we present a novel solution for real-time NMPC
employing a time-mesh refinement strategy. We addressed the planning
and tracking problems in an unified manner by solving a least-squares
optimization formulation, over the flat outputs of the controlled
system.  We evaluated our approach through regulation and tracking
experiments with an UAV simulated platform.  The experiments suggest
that the proposed time-mesh refinement allows to improve the accuracy
of the solution enabling better control performances, without
significantly increasing the computational effort. We release our
C++ open-source implementation, enabling to test the proposed
algorithm with different robotic platforms.  As a future work, we will
investigate additional error constraints over longer horizons.

\appendices
\section{Differential Flatness for UAV}
\label{sec:dynamic_flat_model}

Let $\referenceInertia$ be the right-hand inertial reference frame
with unit vectors along the axes denoted by
$\{\inertialVectorComponent_x, \inertialVectorComponent_y,
\inertialVectorComponent_z\}$. The vector $p = (p_1,p_2,p_3) \in
\referenceInertia$ denotes the position of the center of mass of the
vehicle.

Let $\referenceBody $ be the right-hand body reference frame with unit
vectors $ \{\bodyVectorComponent_x, \bodyVectorComponent_y,
\bodyVectorComponent_z\} $, where these vectors are the axes of frame
$\referenceBody$ with respect to frame $\referenceInertia$.  The
orientation of the rigid body is given by the rotation $\iRb = R
= \begin{bmatrix} \bodyVectorComponent_x & \bodyVectorComponent_y &
  \bodyVectorComponent_z\end{bmatrix} \in \SO(3)$.

Let $\linearVelocity \in \referenceInertia$ express the linear
velocity of the body, expressed in the inertial reference frame
$\referenceInertia$. Let $\angularVelocity \in \referenceBody$ be the
angular velocity of the body with respect to $\referenceInertia$. Let
$\mass$ denote the mass of the rigid body, and $\bI \in \real^{3x3}$
the constant inertia matrix expressed in body frame, the rigid body
equations can be expressed as
\begin{subequations}
\label{eq:dynamic_model}
\begin{align}
  & \dot{\xi} = \mathit{v},\\
  & \mass \dot{v} = \mass \gravity \inertialVectorComponent_z + R F_t\\ 
  & \dot{R} =
  R~\angularVelocity_\times\\ 
  & \bI \dot{\angularVelocity}=-\angularVelocity \times \bI\angularVelocity + \tau
\end{align}
\end{subequations}
where the notation $\angularVelocity_\times$ denotes for the
skew-symmetric matrix formed from $\angularVelocity$. The system
inputs $F_t, \tau \in \referenceBody$ act respectively as thrust force
and body torques.
The system reported in \eqref{eq:dynamic_model} can be
represented exploiting the \textit{differential flatness} property.
For an UAV underactuated vehicle, the flat outputs are given as $\flatoutput = (
p_1, p_2, p_3, \gamma ) \in \real^4 $, where $\gamma$ represents the
yaw angle.  Hence, by denoting $p_t = ( p_1, p_2, p_3)$, we recover
the full state and controls by using the following relations: $p =
p_t$, $\dot{p} = \dot{p_t}$, $F_t = \| \mass(\ddot{p_t} -\gravity) \|$ and
the three columns of the rotation matrix $R$ as:
\begin{eqnarray}
&R_z = \mass(\ddot{p_t} -\gravity) / F_t\nonumber\\ &R_y = R_z
  \times \begin{pmatrix} cos~\mathit{\gamma_t} \\ sin~\mathit{\gamma_t}
    \\ 0 \end{pmatrix} /\Bigg\| R_z \times \begin{pmatrix}
    cos~\mathit{\gamma_t} \\ sin~\mathit{\gamma_t} \\ 0 \end{pmatrix}
  \Bigg\|\nonumber\\ &R_x = R_y \times R_z.\nonumber
\end{eqnarray}
The angular velocity is recovered as
\begin{eqnarray}
& \angularVelocity_x = -R_y~\dddot{p_t}/F_t\nonumber\\ &
  \angularVelocity_y = R_x~\dddot{p_t}/F_t\nonumber\\ &
  \angularVelocity_z =
  \dot{\mathit{\gamma_t}}~(\inertialVectorComponent_z~R_z)\nonumber
\end{eqnarray}
where $\inertialVectorComponent_z$ is the standard unit vector along
the z-axis. To recover $\tau$, we first recover
$\dot{\angularVelocity}$. Note that from the dynamics
\begin{eqnarray}
&\mass~p_t^{(4)} = (R\hat{\angularVelocity}^2 +
  R\hat{\dot{\angularVelocity}})F_t~\inertialVectorComponent_z +
  2~R~\hat{\angularVelocity}\dot{F_t}\inertialVectorComponent_z+R~\ddot{F_t}~\inertialVectorComponent_z\nonumber
\end{eqnarray}
Solving this for $\dot{\angularVelocity}$ gives
\begin{eqnarray}
  & \angularVelocity_x = (-\mass R_y~p_t^{(4)} -
    \angularVelocity_y\angularVelocity_zF_t +
    2\angularVelocity_x\dot{F_t})/F_t\nonumber\\ & \angularVelocity_y =
    (\mass R_y~p_t^{(4)} - \angularVelocity_x\angularVelocity_zF_t -
    2\angularVelocity_y\dot{F_t})/F_t\nonumber\\ & \angularVelocity_z =
    \ddot{\gamma_t}\inertialVectorComponent_z~R_z +
    \dot{\gamma_t}\inertialVectorComponent_z^TR\hat{\angularVelocity}\inertialVectorComponent_z\nonumber
\end{eqnarray}
 Then we use the dynamics $\tau = \bI\dot{\angularVelocity} -
\bI\angularVelocity \times \angularVelocity$ and $\tau$ is recovered.

\balance

\bibliographystyle{plain}
\bibliography{glorified.bib}

\end{document}